# Approximate Inference Algorithms for Hybrid Bayesian Networks with Discrete Constraints

**Vibhav Gogate and Rina Dechter**
Donald Bren School of Information and Computer Science
University of California, Irvine, CA 92967
{vgogate,dechter}@ics.uci.edu

## Abstract

In this paper, we consider *Hybrid Mixed Networks (HMN)* which are Hybrid Bayesian Networks that allow discrete deterministic information to be modeled explicitly in the form of constraints. We present two approximate inference algorithms for HMNs that integrate and adjust well known algorithmic principles such as Generalized Belief Propagation, Rao-Blackwellised Importance Sampling and Constraint Propagation to address the complexity of modeling and reasoning in HMNs. We demonstrate the performance of our approximate inference algorithms on randomly generated HMNs.

## 1 INTRODUCTION

In this paper, we present and evaluate approximate inference algorithms for Hybrid Mixed Networks which are Hybrid Bayesian Networks that contain discrete deterministic information in the form of constraints. Our work is motivated by a real-world application of reasoning about car-travel activity of individuals. This application was modeled using Dynamic Bayesian Networks and requires expressing discrete and continuous variables as well as deterministic discrete constraints.

The two popular approximate inference algorithms used for inference in Dynamic Bayesian Networks (DBN) are Rao-Blackwellised Particle Filtering [Doucet et al., 2000] and Expectation Propagation [Heskes and Zoeter, 2002]. These algorithms use Importance Sampling, exact inference and Generalized Belief Propagation (GBP) on a Bayesian Network, which is a basic structural component of a DBN. We seek to extend these algorithms to our application in which the basic structural component is a Hybrid Mixed Network (HMN).

We show how exact inference algorithms like join-tree clustering and a parameterized GBP algorithm called Iterative Join Graph Propagation (IJGP) can be extended to HMNs in a straightforward way. However, extending Rao-Blackwellised Importance Sampling algorithms (RB-Sampling) to HMNs using the straightforward way results in poor performance. This is because in HMNs every sample that violates a constraint will receive zero weight and will be rejected. To remedy this problem, we suggest a new Importance Sampling algorithm called IJGP-RB-Sampling which uses the output of IJGP as an importance function and we view it as the main contribution of this paper.

We performed experiments on random HMNs to compare how IJGP, pure RB-Sampling and IJGP-RB-Sampling perform relative to each other in terms of accuracy when given the same amount of time. Our empirical results suggest that IJGP-RB-Sampling is always better than pure RB-Sampling and dominates IJGP as the constraint tightness increases.

The rest of the paper is organized as follows. Section 2 defines HMNs and presents some preliminaries. In the two subsequent sections, we describe how to extend join tree clustering and Iterative Join Graph Propagation to HMNs. We follow by describing IJGP-RB-Sampling for performing effective Rao-Blackwellised Importance Sampling in HMNs. We then present empirical results on random HMNs and conclude with a discussion of related work and summary.

## 2 PRELIMINARIES AND DEFINITIONS

A graphical model is defined by a collection of functions, over a set of variables, conveying probabilistic or deterministic information, whose structure is captured by a graph.

DEFINITION **2.1** *A **graphical model** is a triplet $(X,D,F)$ where 1. $X = \{x_1, x_2, \ldots, x_n\}$ is a finite set of variables, 2. $D = \{D_1, \ldots, D_n\}$ is a set of domains of values in which $D_i$ is a domain of $X_i$ and 3. $F = \{F_1, F_2, \ldots, F_m\}$ is a set of real-valued functions. The scope of functions $f_i$ denoted as $scope(f_i) \subseteq X$, is the set of arguments of $f_i$.*

DEFINITION 2.2 *The primal graph of a graphical model is an undirected graph that has variables as its vertices and an edge connects any two variables that appear in the scope of the same function.*

Two graphical models of interest in this paper are Hybrid Bayesian Networks and Constraint Networks. A **Hybrid Bayesian Network (HBN)** [Lauritzen, 1992] $\mathcal{B} = (X, D, P)$ is defined over a directed acyclic graph $G = (X, E)$ and its functions $P_i = \{P(x_i|pa_i)\}$ where $pa_i$ is the set of *parent* nodes of $x_i$. $X$ is the set of variables partitioned into discrete $\Delta$ and continuous $\Gamma$ variables, i.e. $X = \Gamma \bigcup \Delta$. The graph structure $G$ is restricted in that continuous variables cannot have discrete variables as their child nodes. The conditional distribution of continuous variables are given by a linear Gaussian model: $P(x_i|I=i, Z=z) = N(\alpha(i) + \beta(i)*z, \gamma(i))$ $x_i \in \Gamma$ where $Z$ and $I$ are the set of continuous and discrete parents of $x_i$ respectively and $N(\mu, \sigma)$ is a multi-variate normal distribution. The network represents a joint distribution over all its variables given by a product of all its CPDs. A **Constraint network** [Dechter, 2003] $\mathcal{R} = (X, D, C)$, has $C = \{C_1, C_2, \ldots, C_m\}$ as its functions also called as constraints. Each constraint $C_i$ is a relation $R_i$ defined over a subset of the variables $S_i \subseteq X$ and denotes the combination of values that can be assigned simultaneously. A *Solution* is an assignment of values to all the variables such that no constraint is violated. The primary query is to decide if the Constraint Network is consistent (whether it has a solution) and if so find one or all solutions.

Using the Mixed Network framework [Dechter and Mateescu, 2004] for augmenting Bayesian Networks with constraints, we can extend HBNs to include discrete constraints, yielding *Hybrid Mixed Networks (HMNs)*. Formally,

DEFINITION 2.3 (**Hybrid Mixed Network**) *Given a HBN $\mathcal{B} = (X, D, P)$ that expresses the joint probability $P_\mathcal{B}$ and given a Constraint Network $\mathcal{R} = (\Delta, D_\Delta, C)$ that expresses a set of solutions $\rho$, a Hybrid Mixed Network (HMN) based on B and R denoted by $M(B, R)$ is created from $\mathcal{B}$ and $\mathcal{R}$ as follows. The discrete variables $\Delta$ and their domains are shared and the relationships include the CPDs in P and the constraints in C. We assume that the Constraint Network is always consistent and so the HMN expresses the conditional probability $PM(X)$: $P_M(x) = P_\mathcal{B}(x|x \in \rho)$ if $x \in \rho$ and 0 otherwise.*

**Example 2.1** *Figure 1 shows a HBN and a Constraint Network yielding a HMN over variables $\{A,B,C,D,F,G\}$ where D and G are continuous variables (drawn as squares) and the rest are discrete (drawn as circles).*

DEFINITION 2.4 (**Graph Decomposition**) *Given a graphical model $(X, D, F)$, a graph decomposition is a triplet $(GD, \chi, \psi)$ where $GD(V, E)$ is a graph and $\chi$ and $\psi$ are*

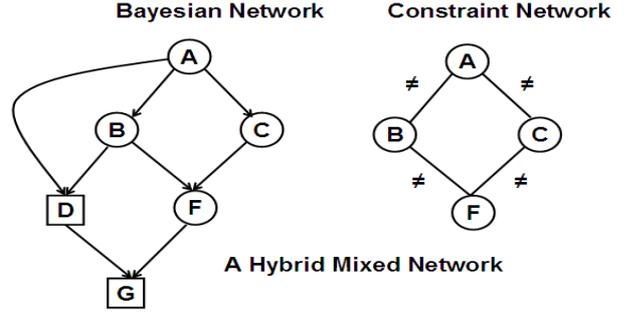

Figure 1: Example HMN consisting of a HBN and a Constraint Network

*labeling functions which associate with each vertex $v \in V$ two sets, $\chi(v) \subseteq X$ and $\psi(v) \subseteq F$ such that, (1) For each function $f_i \in F$, there is exactly one vertex $v \in V$ such that $f_i \in \psi(v)$, and $scope(f_i) \subseteq \chi(v)$, (2) For each variable $x_i \in X$, the set of $\{v \in V | x_i \in \chi(v)\}$ induces a connected subgraph of G (called the running intersection property).*

The *width* of a graph-decomposition is $w = max|\chi(v)| - 1$. A *join-tree-decomposition* is a graph-decomposition in which the graph is a tree while *join-graph-decompositions* $(JG(i))$ are graph-decompositions in which the width is bounded by $i$.

**Example 2.2** *Figure 2 showing (a) primal graph, (b) join-graph-decomposition and (c) join-tree-decomposition of the example HMN shown in Figure 1.*

Another relevant notion is that of *w-cutset*. Given a graphical model $(X, D, F)$, the *w-cutset* is the set of variables $X_1 \subseteq X$ whose removal from the graphical model yields a graphical model whose treewidth is bounded by $w$.

This paper focuses on the problem of computing the posterior marginal distribution (or beliefs) at each variable given evidence i.e. $P(x_i|e)$. This problem is known to be NP-hard and so we resort to approximations.

## 3 EXACT INFERENCE IN HMNs

In this section, we extend a class of exact inference algorithms based on join-tree-clustering [Lauritzen, 1992] from HBNs to HMNs. This algorithm will serve as a basis for the Generalized Belief Propagation scheme described in section 4 which will be investigated empirically as a stand-alone scheme and also as a component in our Importance Sampling scheme described in section 5.

The join-tree-clustering algorithm for HMNs can be derived in a straightforward way by incorporating ideas from [Lauritzen, 1992] and we describe it here for completeness sake (see Figure 3). The exact inference algorithm in [Lauritzen, 1992] works by first forming a join-tree-decomposition and then passing messages between individual cliques of a join-tree-decomposition. A message from node $N_i$ to $N_j$ is constructed by first multiplying all messages and functions in $N_i$ excluding the message from $N_j$ and then marginalizing the product over the separator between $N_i$ and $N_j$. The operators of marginalization $\Downarrow$ and multiplication $\otimes$ required for message-passing on a join-tree-decomposition of a HMN can be constructed in a straightforward way by combining the operators in [Lauritzen, 1992] and [Dechter, 2003] that work on HBNs and constraint relations respectively. We will now briefly comment on how the multiplication operator can be derived. Let us assume we want to multiply a collection of probabilistic functions $P'$ and a set of constraint relations $C'$ (which consist of discrete tuples allowed by the constraint)
to form a single function $PC$. Here, multiplication can be performed on the functions in $P'$ and $C'$ separately using the operators in [Lauritzen, 1992] and [Dechter, 2003] respectively to compute a single probabilistic function $P$ and a single constraint relation $C$. These two functions $P$ and $C$ can be multiplied by deleting all tuples in $P$ that are not present in $C$ to form the required function $PC$.

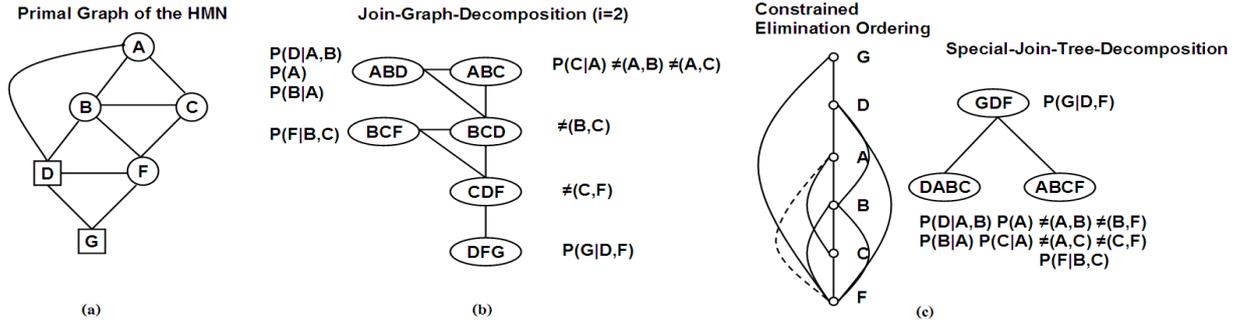

Figure 2: Graph decompositions of HMN in Figure 1.

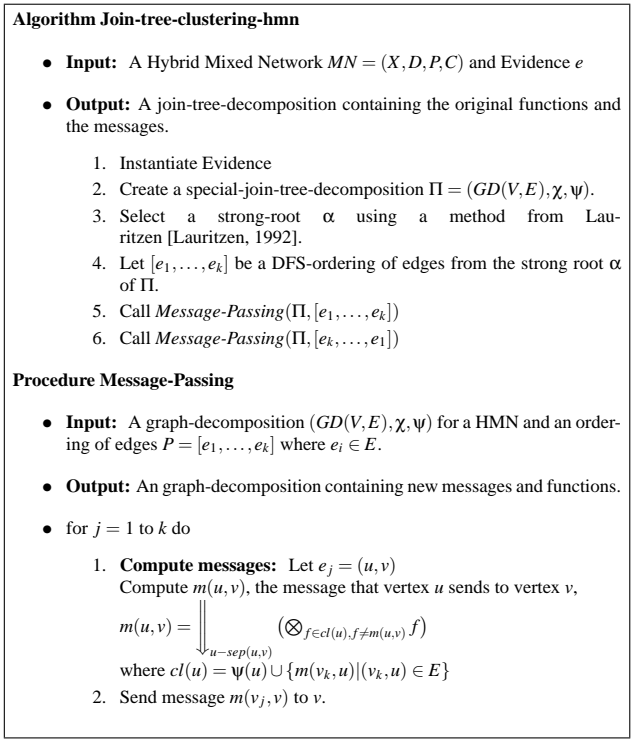

Figure 3: Join-tree-clustering for HMNs

We comment on two major technical points for the algorithm given in Figure 3. Firstly, to be sound the join-tree-clustering algorithm must satisfy the strong root property as required by HBNs [Lauritzen, 1992]:

DEFINITION 3.1 (**Strong Root**) *Given a join-tree-decomposition $(GD(V,E),\chi,\psi)$, a node $r \in V$ is a **strong-root** iff for all neighboring nodes $c$ and $d$ with $c$ closer to $r$ than $d$, we have that $sep(c,d) \subseteq \Delta$ or $d \setminus c \subseteq \Gamma$ where $\Gamma$ is the set of continuous variables and $\Delta$ is the set of discrete variables.*

A sufficient condition to ensure that there is at least one strong root in a join-tree-decomposition is to use an ordering for triangulation in which all continuous variables are ordered before discrete variables. We call such join-tree-decompositions as *special-join-tree-decompositions* (see Figure 2(c)).

Finally, because Gaussian nodes can be processed in polynomial time, the complexity of processing each clique is exponential only in the number of discrete variables in the clique. We capture this property using the definition of adjusted-width.

DEFINITION 3.2 *Given a join-graph-decomposition $(GD,\chi,\psi)$, the **adjusted-width** of a join-graph decomposition is $w = \max |\chi(v) \cap \Delta| - 1$. The adjusted-treewidth of join-tree-decomposition is equal to its adjusted-width.*

It is straight forward to show that [Lauritzen, 1992]:

THEOREM 3.1 *Given a HMN $MN(X,D,P,C)$ and evidence $e$, algorithm Join-tree-clustering-hmn is sound. For discrete variables, the marginal at each clique computed by multiplying the messages and functions in each clique is*

*exact while for continuous variables the marginal is exact in the sense that it has the correct first and second moments as the exact marginal.*

THEOREM 3.2 *The time-complexity of Join-tree-clustering-hmn is $O(|\Delta| * |\Gamma_c|^3 * d^{w*+1}))$. Here $\Gamma_c$ is the maximum number of continuous variables in the clique of a join-tree-decomposition, d is the maximum domain size of the discrete variables, $\Delta$ is the set of discrete variables in the HMN and w∗ is the adjusted-treewidth of the special-join-tree-decomposition used.*

## 4 ITERATIVE JOIN GRAPH PROPAGATION

In this section, we extend an approximate inference algorithm called Iterative Join Graph Propagation (IJGP) to HMNs. IJGP(i) [Dechter et al., 2002] is parameterized Generalized Belief Propagation algorithm which operates on a join-graph-decomposition having less than $i + 1$ variables in each clique. The complexity of IJGP(i) is bounded exponentially by $i$, also called the $i$-bound. This algorithm was defined for discrete Bayesian Networks in [Dechter et al., 2002].

IJGP(i) can be extended to HMNs in a straight-forward way by iteratively applying the message-passing procedure given in Figure 3 to a join-graph-decomposition until a maximum number of iterations is performed or until the algorithm converges.

An important technical difference between the extension of IJGP(i) to HMNs and the original IJGP(i) algorithm [Dechter et al., 2002] is that $i$ stands for adjusted-width rather than width.

THEOREM 4.1 *The complexity of IJGP(i) when applied to HMN is $O((|\Delta| + n) * d^i * |\Gamma_c|^3)$ where $|\Delta|$ is the number of discrete variables, d is the maximum-domain size of the discrete variables, i is the adjusted-i-bound, n is the number of nodes in a join-graph and $|\Gamma_c|$ is the maximum number of continuous variables in any clique of the join-tree-decomposition used.*

## 5 RAO-BLACKWELLISED IMPORTANCE SAMPLING

In this section, we propose an effective Importance Sampling for HMNs. We will first review Importance Sampling algorithms for computing posterior distribution and then review w-cutset sampling which is a special version of the Rao-Blackwellisation concept. Subsequently, we discuss how an Importance Sampling algorithm would run into problems when hard constraints are present. We end the section by presenting an algorithm called IJGP-RB-Sampling that remedies these problems by using Iterative Join Graph Propagation to generate an effective proposal distribution.

Sampling methods are used for approximate inference in Bayesian Networks and are useful in cases when the distribution is hard to compute analytically using exact inference. The virtue of sampling schemes is that they are guaranteed to yield the correct posterior distribution when they converge and they use only linear space. An important class of sampling algorithms is Importance Sampling for Bayesian Networks. The idea here is that since we cannot sample from the true posterior $P(X|e)$ (while it is NP-hard to compute), we will sample from an approximation $Q(X)$ such that the ratio $w = \beta * P(X = x|e)/Q(X = x)$[1] is known up to a normalizing constant $\beta$. We can then compute the required posterior marginal as $P(x_i = x|e) = \sum_j f_j(x) * \widetilde{w_j}$ where $f_j(x)$ is the sample that agrees with $x_i = x$ and $\widetilde{w_j}$ are the normalized importance weights computed as $\widetilde{w_j} = w_j / \sum_k w_k$. Ideally, the proposal distribution should have the following properties: (1) It is easy to sample from (2) It allows easy evaluation of the value $Q(X = x)$ for each sample so that the weights can be computed in a cost-effective manner and (3) If $P(X) \neq 0$ then $Q(X) \neq 0$. The last property ensures that Importance Sampling converges to the true posterior in limit of convergence [Geweke, 1989].

It is well known that any sampling scheme over multi-dimensional space can be assisted by Rao-Blackwellised sampling, namely by sampling over a subspace. We now describe w-cutset sampling which is a special version of Rao-Blackwellised (RB) sampling. w-cutset sampling [Bidyuk and Dechter, 2003] is a method that combines exact inference and sampling and provides a systematic scheme for sampling from a subset of variables. The idea is that given an assignment to a set of variables it might be possible to compute the remaining distribution analytically. More formally, in w-cutset sampling we partition the set of variables $X$ into two subsets $X = X_1 \cup X_2$ such that the treewidth of the graphical model when $X_1$ is removed is bounded by $w$. Each w-cutset sample consists of an assignment of values to $X_1 = x_1$ and a belief state $P(X_2|x_1)$. The variables in the set $X_1$ are sampled and the remaining $X_2$ variables are solved exactly using exact algorithms like join-tree-clustering.

We can straightforwardly adapt w-cutset Importance Sampling to Hybrid Bayesian Networks (HBNs). Since exact inference is polynomial if all nodes are Gaussian, w-cutset sampling in HBNs can be done by sampling only a subset of the discrete variables [Lerner, 2002]. Extending this idea to HMNs, suggests that we sample the discrete variables using a suitable proposal distribution and discard all samples that violate one or more constraints. This method can be inefficient. For example, if we use the prior as the proposal distribution (as in Likelihood weighting) and

---
[1] This is usually called Biased Importance Sampling

```
Ordered Buckets structure (Ordering: [A B C F])

A:    P(A)

B:    P(B|A)  m(1,2)(AB)  m(2,1)(AB)  ≠(A,B)

C:    m(2,4)(BC)  m(4,2)(BC)  m(3,4)(BC)  m(4,3)(BC)  P(C|A)  ≠(A,C)
      ≠(B,C)

F:    m(3,5)(CF)  m(5,3)(CF)  ≠(C,F)  P(F|B,C)
```

Figure 4: An ordered Buckets structure for the join-graph-decomposition in Figure 2. m(x,y) is the message sent by node x to node y.

the prior is such that solutions to the constraint portion are highly unlikely, a large number of samples will be rejected (because $P(X^i = 0)$ for a sample $X^i$ and so weight would be 0).

On the other hand, if we want to make the sample rejection rate zero we would have to use a proposal distribution $Q$ such that all samples from $Q$ are solutions of the constraint portion. One way to find this proposal distribution is to make the Constraint Network backtrack-free (perhaps using adaptive-consistency [Dechter, 2003]) along an ordering of variables and then sample along a reverse ordering. However, adaptive-consistency can be costly unless the treewidth of the constraint portion is small. Thus on one hand, zero-rejection rate implies using a costly inference procedure and on the other hand, sampling from a proposal distribution that ignores the constraints may result in a high rejection rate.

We propose to exploit the middle ground between the two extremes by combining the Constraint Network and the Bayesian Network into a single approximate distribution $Q$ using IJGP(i). By using IJGP(i) we are likely to reduce the rejection-rate because it applies constraint-propagation in the form of relational i-consistency [Dechter and Mateescu, 2003], namely it removes many inconsistent tuples [Dechter, 2003]. Note that the output of IJGP(i) can be used to generate a proposal distribution because as shown in [Dechter and Mateescu, 2003] $P(X|e) > 0$ implies that $Q(X|e) > 0$ where $Q(X|e)$ is the distribution of IJGP(i).

We now describe a method to generate samples from the output of IJGP(i). Here, given an ordering $\pi = \langle x_1, \ldots, x_j \rangle$ of the discrete variables to be sampled, we first compute an approximate marginal denoted by $Q(x_1)$ from the output of IJGP($i$) and then sample $x_1$ from $Q(x_1)$. Then, we set the sampled value $x_1 = a_1$ as evidence, run IJGP(i), compute the marginal $Q(x_2|x_1 = a_1)$ and sample $x_2$ from this marginal. The above process is repeated until all variables are sampled. The method is inefficient however, requiring $O(|X_1| * exp(i) * N * d)$ time for generating all samples

```
Algorithm IJGP-RB-Sampling

• Input: A Hybrid Mixed Network MN(X,D,P,C) and Evidence e. Integer i, k, w
  and N.

• Output: Estimate of P(X|e).

• Perform Iterative Join-graph propagation on MN with i-bound=i and number of
  iterations=k. Let us call its output Π.

• Partition the Variables of HMN into X₁ and X₂ such that the adjusted-treewidth of
  a special- join-tree-decomposition of X₂ is bounded by w.

• Create a bucket-tree BT(V,ψ) from Π such that V contains only variables in X₁.

• For i = 1 to N do
    1. sᵢ = Generate a sample from BT along the order d of BT for the set of
       variables X₁.
    2. Use join-tree-clustering to compute the distribution on X2 by setting evi-
       dence as e ∪ X₁ = sᵢ. Lets call it rᵢ.
    3. Reject the sample if rᵢ is not a solution.
    4. Compute the importance weights wᵢ of sᵢ.

• Normalize the importance weights wᵢ.

• Output the samples [sᵢ,rᵢ] and the normalized weights wᵢ
```

Figure 5: IJGP-RB-Sampling for Hybrid Mixed Networks

where $X_1$ are the sampled variables, $N$ is the number of samples, $i$ is the i-bound used and $d$ is the maximum domain size.

Instead, we use a simplified method in which IJGP(i) is applied just once yielding a time-complexity of $O(exp(i) + N * |X_1| * d)$ to generate all samples. The simplified method uses a special data-structure of ordered buckets. Given a collection of functions and messages as the output of IJGP(i) and an ordering $\pi = \langle x_1, \ldots, x_j \rangle$ of the discrete variables to be sampled, we construct the ordered buckets structure as follows. We associate a bucket with each variable $x_i$ in $\pi$ and consider only those functions and messages, $F_\pi$ whose scope is included in $\{x_j, \ldots, x_1\}$. We then start processing from i=j to 1 putting all functions in $F_\pi$ that mention $x_i$ in the bucket of $x_i$. Once the ordered buckets structure is created, we sample along the order from $i = 1$ to $j$. The construction procedure guarantees that when we sample a variable $x_i$ from its bucket, all variables ordered before $x_i$ are instantiated and there is only a single un-instantiated variable in each function in the bucket of $x_i$. So, the time-complexity to sample each bucket is bounded by $O(d)$ yielding a time-complexity of $O(N * |X_1| * d)$ to generate all samples. An example ordered buckets structure for the join-graph-decomposition in Figure 2(b) is given in Figure 4.

We now describe how to compute the weight of each sample. According to Rao-Blackwellised Importance Sampling theory [Geweke, 1989, Doucet et al., 2000], the weight of each sample $X_1^k$ over variables $X_1$ is given by $w_k = P'/Q'$ such that $P'/Q' = \beta * P(X_1^k|e)/Q(X_1^k|e)$, where β is a constant. We can determine the quantity $P(X_1^k, e)$ using join-tree-clustering while we can compute $Q(X_1^k, e)$ (up to a normalizing constant) from the ordered buckets struc-ture described above by multiplying individual probabilities. Now since $Q(e) = 1$, we have $\frac{P(X_1^k, e)}{Q(X_1^k, e)} = \beta * \frac{P(X_1^k|e)}{Q(X_1^k|e)}$, as required.

An important advantage of using IJGP(i) in addition to constraint-propagation is that it may yield good approximation to the true posterior thereby proving to be an ideal candidate for proposal distribution. The integration of the ideas expressed above into a formal algorithm called IJGP-RB-sampling is given in Figure 5. The algorithm first runs IJGP(i) for $k$ iterations to generate an approximation to the true posterior. Then, it partitions the variables $X$ into two sets $X_1$ and $X_2$ such that the treewidth of the special join-tree-decomposition of $X_2$ is bounded by $w$ using a method proposed in [Bidyuk and Dechter, 2004]. It then creates an ordered bucket structure over $X_1$ from the output of IJGP(i) and performs Importance Sampling using the ordered buckets structure as described above. We conclude that:

THEOREM **5.1** *The complexity of IJGP-RB-Sampling(i,w) is $O([N * d^{w+1} * |\Gamma_c|^3 * |\Delta|] + [(|\Delta| + n) * d^i * |\Gamma_c|^3])$ where $\Delta$ is the set of discrete variables, d is the maximum-domain size, i is the adjusted-i-bound, w is the adjusted-w-cutset, n is the number of nodes in the join-graph and $|\Gamma_c|$ is the maximum number of continuous variables in the clique of the join-graph-decomposition.*

## 6 EXPERIMENTAL EVALUATION

We tested the performance of IJGP(i), pure RB-Sampling and IJGP-RB-Sampling(i,w) on randomly generated HMNs. We used a parametric model $(N_1, N_2, K, C_1, C_2, P, T)$ where $N_1$ is the number of discrete variables, $N_2$ is the number of Gaussian Variables, $K$ is the domain-size for each discrete variable, $C_1$ is the number of constraints allowed and $T$ is the tightness or the number of forbidden tuples in each constraint, $C_2$ is the number of conditional probability distributions (CPDs) and $P$ is the number of parents in each CPD. Parents in each CPD are picked randomly and each CPD is filled randomly. Note that each Gaussian CPDs was assigned a mean and variance randomly chosen in the range $(0, 1)$. Also no Gaussian variables have discrete children in our random problems. The constraint portion is generated according to Model B [Smith, 1994]. In Model B, for a given $N_1$ and $K$, we select $C_1$ constraints uniformly at random from the available $N(N-1)^2$ binary constraints and then for each constraint we select exactly $T$ tuples (called as constraint tightness) as no-goods (or forbidden) from the available $K^2$ tuples.

We generated two classes of problems (a) a 50-variable set with parameters $(40, 10, 4, 80, 35, 3, T)$ and $T$ was varied with values 4, 6 and 8 and (b) a 100-variable-set with parameters $(90, 10, 4, 180, 95, 3, T)$ and $T$ was varied with values 4, 6 and 8. In each problem class, 10% of the variables were randomly selected as evidence variables. Each algorithm was given the same amount of time for computing approximate posterior Beliefs. For the 50-variable-set, we let each algorithm run for 20s while for the 100-variable-set we let each algorithm run for 100s. The choice of these time-bounds was arbitrary. Also for each IJGP-RB-Sampling(i,w) algorithm instance IJGP(i) is run for 10 iterations only.

For each network, we compute the exact solution using the join-tree-clustering algorithm and compare the accuracy of algorithms using: 1. *Absolute error* - the absolute value of the difference between the approximate and the exact, averaged over all values, all variables and all problems. 2. *Relative error* - the absolute value of the difference between the approximate and the exact, divided by the exact, averaged over all values, all variables and all problems. 3. *KL distance* - $P_e(x_i) * log(P_e(x_i)/P_a(x_i))$ averaged over all values, all variables and all problems where $P_e$ and $P_a$ are the exact and approximate probability values for variable $x_i$ respectively.

For IJGP(i), we experimented with $i$-bounds of 2, 4 and 6 while for IJGP-RB-Sampling $(i, w)$, we experimented with $i$-bound and $w$ of 2, 4 and 6 each. We also experimented with a $w$-cutset Importance Sampling algorithm (or pure RB-Sampling) with $w$ being set to 0, 2, 4 and 6. Thus, we have a total of 16 algorithms in our experimental setup. We tabulate the results using a 4x4 matrix for each combination of the problem-set, value of tightness $T$ and accuracy-scheme (KL-distance, relative and approximate error). The rows of the matrix are labeled from w=0 to w=6 in increments of 2 corresponding to the $w$ values used while the columns are labeled from $i = 0$ to $i = 6$ corresponding to the $i$-bound used. Note that the column-vector $i = 0$ gives the results for $w$-cutset sampling while the row vector $w = 0$ gives results for IJGP$(i)$ (except for $i = 0$ when it gives results for $w$-cutset sampling). The rest of the matrix contains results for IJGP-RB-Sampling for different values of $i$ and $w$ (see Tables 1 and 2).

### 6.1 EXPERIMENTS ON THE 50-VARIABLE-SET

Results on the 50-variable-set are given in Table 1. The results are averaged over 100 instances each. Here, we see that IJGP(i) has slightly better accuracy than IJGP-RB-Sampling when the problem tightness is low $(T = 4)$ (see Figure 5, Table 1). However, as we increase the tightness to $(T = 8)$ the performance of IJGP(i) is worse than IJGP-RB-Sampling (see Figure 6, Table 1). As expected the performance of $w$-cutset sampling improves as $w$ is increased. However IJGP-RB-sampling shows only a slight improvement in accuracy with increase in $w$. The accuracy of $w$-cutset sampling is always worse than IJGP(i) and IJGP-RB-Sampling and also it deteriorates more rapidly as the tightness is increased (see Table 1).

Table 1: Table showing absolute error, relative error and K-L distance for 50-variable-set.

| T | | Relative Error | | | | Absolute Error | | | | K-L distance | | | |
|---|---|---|---|---|---|---|---|---|---|---|---|---|---|
| | | i=0 | i=2 | i=4 | i=6 | i=0 | i=2 | i=4 | i=6 | i=0 | i=2 | i=4 | i=6 |
| | w=0 | 0.03123 | 0.00746 | 0.00727 | 0.00709 | 0.00772 | 0.00184 | 0.00177 | 0.00164 | 0.00062 | 0.00013 | 0.00012 | 0.00012 |
| | w=2 | 0.02124 | 0.00872 | 0.00823 | 0.00737 | 0.00503 | 0.00213 | 0.00198 | 0.00178 | 0.00042 | 0.00017 | 0.00016 | 0.00011 |
| 4 | w=4 | 0.01782 | 0.00843 | 0.00757 | 0.00934 | 0.00439 | 0.00195 | 0.00173 | 0.00209 | 0.00032 | 0.00013 | 0.00014 | 0.00016 |
| | w=6 | 0.01892 | 0.00914 | 0.00803 | 0.00805 | 0.00414 | 0.00208 | 0.00189 | 0.00208 | 0.00037 | 0.00016 | 0.00015 | 0.00016 |
| | w=0 | 0.0569 | 0.01692 | 0.01224 | 0.01329 | 0.01393 | 0.004 | 0.00287 | 0.03023 | 0.00114 | 0.00031 | 0.00024 | 0.00024 |
| | w=2 | 0.05294 | 0.01234 | 0.01123 | 0.01142 | 0.01293 | 0.00301 | 0.00276 | 0.00275 | 0.00104 | 0.00023 | 0.00019 | 0.00022 |
| 6 | w=4 | 0.04543 | 0.01182 | 0.01078 | 0.01234 | 0.01098 | 0.00218 | 0.00248 | 0.00301 | 0.000874 | 0.00021 | 0.00021 | 0.00023 |
| | w=6 | 0.04593 | 0.01221 | 0.01223 | 0.01287 | 0.01103 | 0.00301 | 0.00296 | 0.00309 | 0.00088 | 0.00023 | 0.00024 | 0.00025 |
| | w=0 | 0.10234 | 0.02393 | 0.01872 | 0.01908 | 0.02559 | 0.00598 | 0.00468 | 0.00477 | 0.0020 | 0.00044 | 0.00036 | 0.00036 |
| | w=2 | 0.09029 | 0.01721 | 0.01089 | 0.01056 | 0.02257 | 0.0043 | 0.00272 | 0.00264 | 0.00177 | 0.00034 | 0.00020 | 0.00021 |
| 8 | w=4 | 0.09102 | 0.00927 | 0.01102 | 0.01012 | 0.02276 | 0.00232 | 0.00276 | 0.00253 | 0.0018 | 0.00016 | 0.0002 | 0.00016 |
| | w=6 | 0.07928 | 0.01023 | 0.01394 | 0.01234 | 0.01982 | 0.00256 | 0.00349 | 0.00309 | 0.0016 | 0.00016 | 0.00026 | 0.00022 |

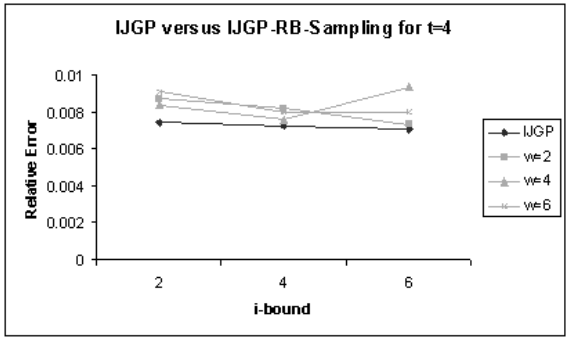

Figure 6: Figure comparing relative error of IJGP and IJGP-RB-Sampling (i,w) for T=4 for 50-variable set

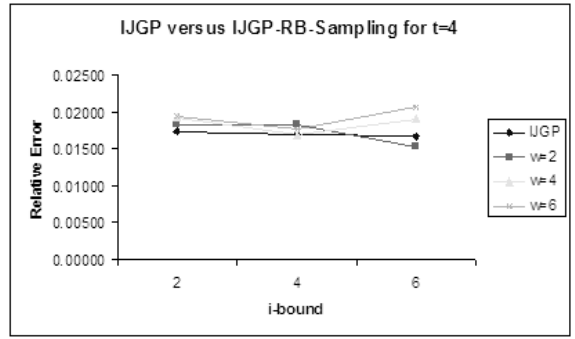

Figure 8: Figure comparing relative error of IJGP and IJGP-RB-Sampling (i,w) for T=4 for 100-variable set

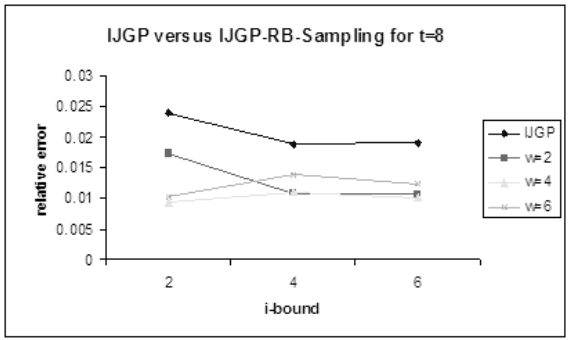

Figure 7: Figure comparing relative error of IJGP and IJGP-RB-Sampling (i,w) for T=8 for 50-variable set

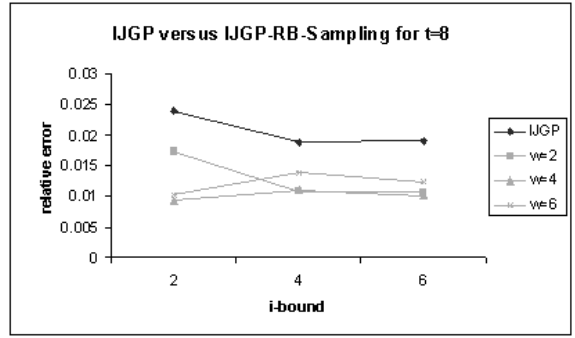

Figure 9: Figure comparing relative error of IJGP and IJGP-RB-Sampling (i,w) for T=8 for 100-variable set

### 6.2 EXPERIMENTS ON THE 100-VARIABLE-SET

Results on the 100-variable-set are given in Table 2. The results are averaged over 100 instances each. Here, we see that unlike the 50-variable-set, IJGP(i) has comparable accuracy to IJGP-RB-Sampling when the tightness is low ($T = 4$) (see Figure 7 and Table 2). However, as we increase tightness ($T = 8$), the accuracy of IJGP(i) is considerably worse than IJGP-RB-Sampling (see Figures 7, 8 and Table 2). Also RB-Sampling is significantly worse than IJGP-RB-Sampling for various values of $w$ (see Table 2).

## 7 RELATED WORK AND SUMMARY

A Mixed Network framework for representing deterministic and uncertain information was presented in [Larkin and Dechter, 2003, Dechter and Mateescu, 2004]. These previous works also describe exact inference algorithms for Mixed Networks with the restriction that all variables should be discrete. Our work goes beyond these previous works in that we describe approximate inference algorithms for the Mixed Network framework and allow continuous Gaussian nodes.

Table 2: Table showing absolute error, relative error and K-L distance for 100-variable-set.

| T | | Relative Error | | | | Absolute Error | | | | K-L distance | | | |
|---|---|---|---|---|---|---|---|---|---|---|---|---|---|
| | | i=0 | i=2 | i=4 | i=6 | i=0 | i=2 | i=4 | i=6 | i=0 | i=2 | i=4 | i=6 |
| | w=0 | 0.06676 | 0.01734 | 0.01692 | 0.01684 | 0.01487 | 0.00369 | 0.00320 | 0.00318 | 0.00128 | 0.00025 | 0.00024 | 0.00023 |
| | w=2 | 0.04220 | 0.01841 | 0.01829 | 0.01542 | 0.01117 | 0.00517 | 0.00414 | 0.00350 | 0.00094 | 0.00039 | 0.00037 | 0.00025 |
| 4 | w=4 | 0.04055 | 0.01926 | 0.01697 | 0.01911 | 0.00937 | 0.00429 | 0.00393 | 0.00428 | 0.00071 | 0.00026 | 0.00032 | 0.00041 |
| | w=6 | 0.03756 | 0.01942 | 0.01765 | 0.02069 | 0.00916 | 0.00431 | 0.00455 | 0.00490 | 0.00083 | 0.00036 | 0.00036 | 0.00039 |
| | w=0 | 0.11526 | 0.03369 | 0.02629 | 0.02136 | 0.03103 | 0.00956 | 0.00609 | 0.06205 | 0.00254 | 0.00064 | 0.00053 | 0.00047 |
| | w=2 | 0.10788 | 0.02658 | 0.02291 | 0.02467 | 0.02913 | 0.00713 | 0.00600 | 0.00577 | 0.00220 | 0.00046 | 0.00039 | 0.00046 |
| 6 | w=4 | 0.10970 | 0.02333 | 0.02468 | 0.02632 | 0.02431 | 0.00512 | 0.00571 | 0.00706 | 0.00203 | 0.00042 | 0.00043 | 0.00047 |
| | w=6 | 0.10043 | 0.02799 | 0.02848 | 0.02889 | 0.02192 | 0.00668 | 0.00621 | 0.00646 | 0.00204 | 0.00047 | 0.00054 | 0.00055 |
| | w=0 | 0.22601 | 0.04838 | 0.04366 | 0.04342 | 0.05453 | 0.01416 | 0.01023 | 0.01105 | 0.00477 | 0.00098 | 0.00081 | 0.00082 |
| | w=2 | 0.18253 | 0.03674 | 0.02384 | 0.02183 | 0.04509 | 0.01026 | 0.00574 | 0.00624 | 0.00386 | 0.00081 | 0.00046 | 0.00048 |
| 8 | w=4 | 0.19833 | 0.01964 | 0.02253 | 0.02295 | 0.04555 | 0.00527 | 0.00645 | 0.00601 | 0.00355 | 0.00035 | 0.00039 | 0.00033 |
| | w=6 | 0.15385 | 0.02392 | 0.03256 | 0.02902 | 0.03890 | 0.00509 | 0.00762 | 0.00633 | 0.00311 | 0.00036 | 0.00061 | 0.00050 |

A class of approximate inference algorithms called IJGP(i) described in [Dechter et al., 2002] handles only discrete variables. In our work, we extend IJGP(i) to include Gaussian variables and discrete constraints.

Importance Sampling is a commonly used algorithm for sampling in Bayesian Networks [Geweke, 1989]. A main step in Importance Sampling is choosing a proposal distribution that is as close as possible to the target distribution. We show how a bounded inference procedure like IJGP(i) can be used to select a good proposal distribution.

The main algorithmic contribution of this paper is presenting a class of Rao-Blackwellised Importance Sampling algorithms, IJGP-RB-Sampling for HMNs which integrates a Generalized Belief Propagation component with a Rao-Blackwellised Importance Sampling scheme for effective sampling in presence of constraints.

Our experimental results are preliminary but very encouraging. Our results on randomly generated HMNs show that IJGP-RB-Sampling is almost always superior to pure *w*-cutset sampling (RB-Sampling) which does not use IJGP as a importance function. Our results also show that IJGP-RB-Sampling has better accuracy than IJGP when the problem tightness is high or when the number of solutions to the constraint portion of HMNs is low.

### ACKNOWLEDGEMENTS

This work was supported in part by the NSF under award numbers 0331707, 0331690 and by NSF grant IIS-0412854.